%% file: main.tex
\documentclass[10pt,twocolumn,letterpaper]{article}
\usepackage{kotex}
\usepackage{graphicx}     
\usepackage[utf8]{inputenc} 

\usepackage{tikz}
\usetikzlibrary{shapes.geometric, arrows.meta, positioning,decorations.pathreplacing, calc}
\usepackage{cvpr}              
\input{preamble}
\definecolor{cvprblue}{rgb}{0.21,0.49,0.74}
\usepackage[pagebackref,breaklinks,colorlinks,allcolors=cvprblue]{hyperref}

\def\paperID{} 
\def\confName{CVPR}
\def\confYear{2026}
 
\title{Point to Span: Zero-Shot Moment Retrieval for Navigating Unseen Hour-Long Videos}

\author{
    Mingyu Jeon\textsuperscript{1},
    Jisoo Yang\textsuperscript{1},
    Sungjin Han\textsuperscript{1},
    Jinkwon Hwang\textsuperscript{1}, \\   
    Sunjae Yoon\textsuperscript{2},
    Jonghee Kim\textsuperscript{3},
    Junyeoung Kim\textsuperscript{1} \\[0.5em] 
    \textsuperscript{1}Department of Artificial Intelligence, Chung-Ang University\\
    \textsuperscript{2}Korea Advanced Institute of Science and Technology (KAIST)\\
    \textsuperscript{3}Electronics and Telecommunications Research Institute (ETRI) \\[0.5em] 
    {\tt\small \{smart2557, junyeongkim\}@cau.ac.kr, sunjae.yoon@kaist.ac.kr, jhkim27@etri.re.kr}
}
\begin{document}

\maketitle

\input{sec/0_abstract}

\input{sec/1_intro}
\input{sec/2_related_works}
\input{sec/3_method}

\input{sec/4_experiments}
\input{sec/5_conclusion}
\input{sec/X_suppl}

\input{sec/_finalcopy}
{
    \small
    \bibliographystyle{ieeenat_fullname}
    \bibliography{main}
}


\end{document}

%% file: sec/0_abstract.tex
\begin{abstract} 
Zero-shot Long Video Moment Retrieval (ZLVMR) is the task of identifying temporal segments in hour-long videos using a natural language query without task-specific training. 
The core technical challenge of LVMR stems from the computational infeasibility of processing entire lengthy videos in a single pass. 
This limitation has established a `Search-then-Refine' approach, where candidates are rapidly narrowed down and only those portions are analyzed, as the dominant paradigm for LVMR.
However, existing approaches to this paradigm face severe limitations. Conventional supervised learning suffers from limited scalability and poor generalization, despite substantial resource consumption. 
Yet, existing zero-shot methods also fail, facing a dual challenge: (1) their heuristic strategies cause a `search' phase candidate explosion, and (2) the ‘refine' phase, which is vulnerable to semantic discrepancy, requires high-cost VLMs for verification, incurring significant computational overhead.
We propose \textbf{P}oint-\textbf{to}-\textbf{S}pan (P2S), a novel training-free framework to overcome this challenge of inefficient `search' and costly `refine' phases. 
P2S overcomes these challenges with two key innovations: an ‘Adaptive Span Generator'  to prevent the search phase candidate explosion, and ‘Query Decomposition' to refine candidates without relying on high-cost VLM verification.
To our knowledge, P2S is the first zero-shot framework capable of temporal grounding in hour-long videos, outperforming supervised state-of-the-art methods by a significant margin (e.g., +3.7\% on R5@0.1 on MAD).
\end{abstract}

%% file: sec/1_intro.tex
\section{Introduction}
\label{sec:intro}

\begin{figure}[htb!]
\centerline{\includegraphics[width=1\columnwidth]{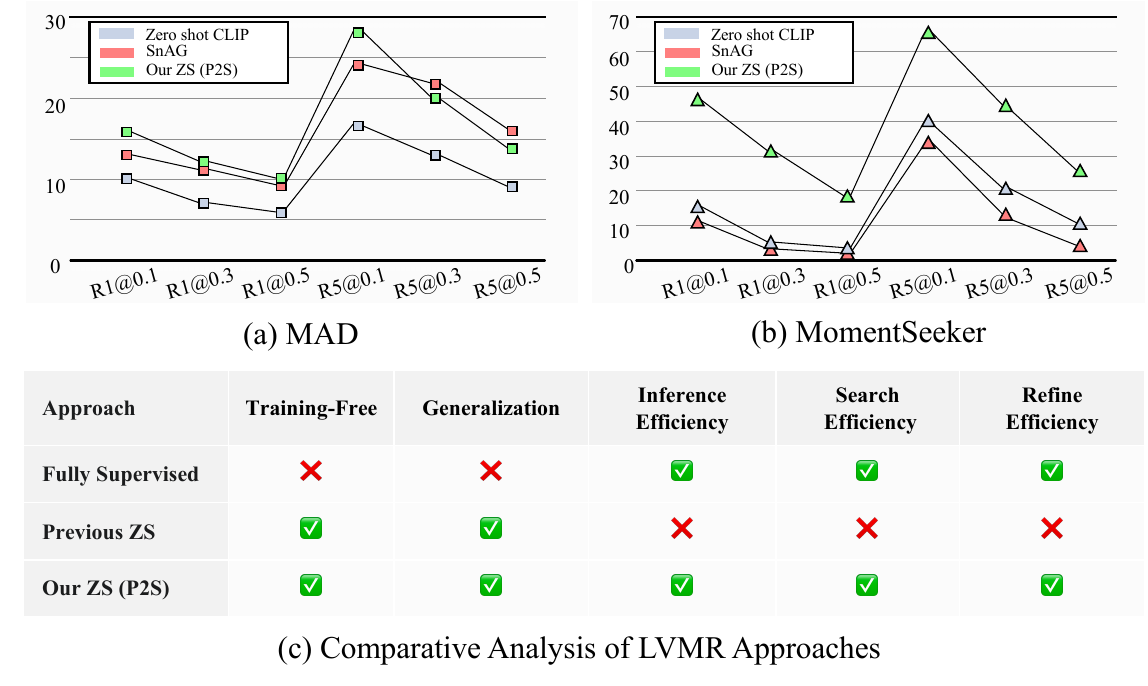}}
\caption{
Comparison of our Zero-shot (ZS) P2S framework with existing approaches.
Quantitative performance against a supervised model (SnAG~\cite{SnAG}), a previous ZS method (Zero-shot CLIP~\cite{CLIP}) and Ours. 
While the supervised model is competitive on its in-distribution dataset (a) MAD, its performance degrades significantly on the out-of-distribution benchmark (b) MomentSeeker, demonstrating its limited generalization.
Our ZS method consistently outperforms both.
(c) A summary illustrating how our framework overcomes the distinct limitations of both paradigms: the poor generalization of supervised models and the poor efficiency of previous ZS methods.
}
\label{fig1}
\end{figure}

The explosive growth of long-form video, spanning from streaming libraries to industrial surveillance, has created a critical need to locate specific moments within archives efficiently. 
Long-form Video Moment Retrieval (LVMR) addresses this need by enabling instant retrieval of temporal segments directly from natural language queries. 

%
%
\begin{figure*}[htb!]
\centerline{\includegraphics[width=1\textwidth]{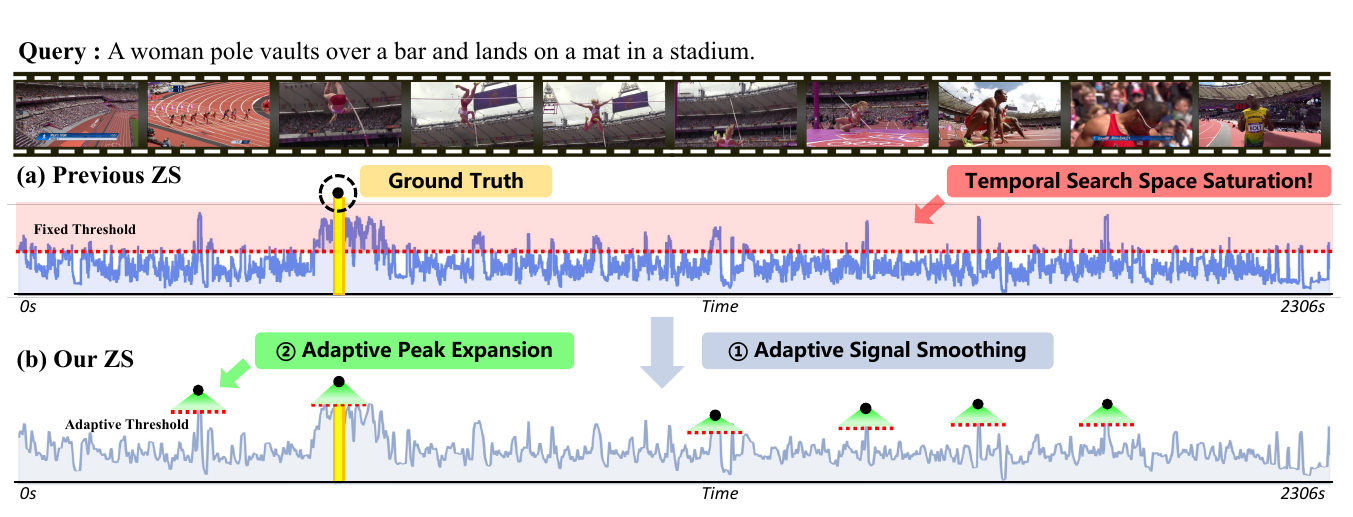}}
\caption{
Illustration of the challenge in long-form videos, which neutralizes the \textit{search} phase. 
(a) Previous ZS methods, designed for short videos, fail to handle the `search space scaling' required for long videos. 
This saturates the search space (red area) with an unmanageable surge of false positives. %
(b) Our ZS method resolves this by dynamically adapting to the signal, robustly identifying only reliable peaks (green spans) and preventing the temporal search space saturation.
} 
\label{fig2}
\end{figure*}

The core technical challenge of LVMR stems from the computational infeasibility of processing such lengthy videos in a single pass.
Modern Vision-Language Models (VLMs) struggle to process extremely long input sequences, making end-to-end understanding across the entire video impractical.
This infeasibility forces the abandonment of holistic understanding in favor of adopting efficient \textit{search} strategies.
Consequently, the prevailing paradigm has shifted toward a \textit{Search-then-Refine} strategy.
This approach first coarsely identifies candidate segments via an efficient query-based search, followed by the precise localization of the target moment within this reduced set of promising regions.

Despite its prevalence, the \textit{Search-then-Refine} approaches~\cite{zvmr_diwan, CONE, SnAG, RevisionLLM} still suffer from significant practical and scalability barriers.
In this paper, we propose Point-to-Span (P2S), a novel training-free framework designed to effectively overcome these limitations. 
The first major barrier lies in conventional supervised learning~\cite{VLG-NET, CONE, SnAG, RevisionLLM}, the mainstream approach, which demands extensive labeled data and long training times. 
Even after such heavy supervision, such models tend to become biased toward specific domains and linguistic styles, resulting in significant performance degradation in novel environments unseen during training (see Figure~\ref{fig1} (a), (b)).
The high cost and poor generalization of supervised paradigms highlight the necessity of a Zero-Shot (ZS) alternative for scalable deployment.

However, existing ZS frameworks~\cite{zvmr_diwan, trainingfree, Moment-GPT,slowfast-llava} remain far from achieving this vision. 
As illustrated in Figure~\ref{fig2}, methods originally designed for short videos fail to handle the temporal search space scaling required by long-form content.
Dense scanning strategies~\cite{Charades-STA} over hour-long sequences lead to \textbf{temporal search space saturation}, characterized by an explosion of false positives that effectively neutralizes the \textit{search} phase. 
The neutralization of the \textit{search} phase forces a severe computational dilemma on the subsequent \textit{refine} phase: precise analysis based on high-cost VLMs~\cite{Moment-GPT, zvmr-Gong, vtimellm, TimeChat, LITA, HawkEye,timesearch,videomind,simplellm,momentor} becomes computationally infeasible, while opting for lightweight similarity-based methods~\cite{zvmr_diwan, trainingfree} remain efficient but fail to capture the full semantic extent of events and leads to \textbf{semantic discrepancy}.
Figure~\ref{fig1} (c) summarizes this dual bottleneck - limited generalization in supervised methods and inefficient search in prior zero-shot approaches.

To address these fundamental limitations, P2S introduces two key technical innovations:
(1) \textbf{Adaptive Span Generator} that dynamically adjusts to the statistical characteristics of the input video, robustly identifying candidate segments even in highly noisy, hour-long videos, and (2) \textbf{Query Decomposition} that converts a natural language query into temporally ordered sub-queries, capturing the inherent compositional structure of multi-step events~\cite{deco, zvmr-Gong, query-dependent}. These sub-queries serve as semantic anchors for the \textit{refine} phase, guiding candidate refinement for better alignment with the actual temporal boundaries of the target event, without any additional model training.
%
%
%
To our knowledge, P2S is the first zero-shot framework capable of temporal grounding in hour-long videos, outperforming high-cost state-of-the-art supervised models~\cite{RevisionLLM, SnAG, SOONet, RGNet} by a significant margin (e.g., +3.7\% on R5@0.1 on MAD~\cite{MAD}). 
Considering the absence of training costs and superior scalability, P2S sets a new standard for practical, generalizable, and computationally efficient LVMR. 
%





%% file: sec/2_related_works.tex
\begin{figure*}[htb!]
\centerline{\includegraphics[width=1.0\textwidth]{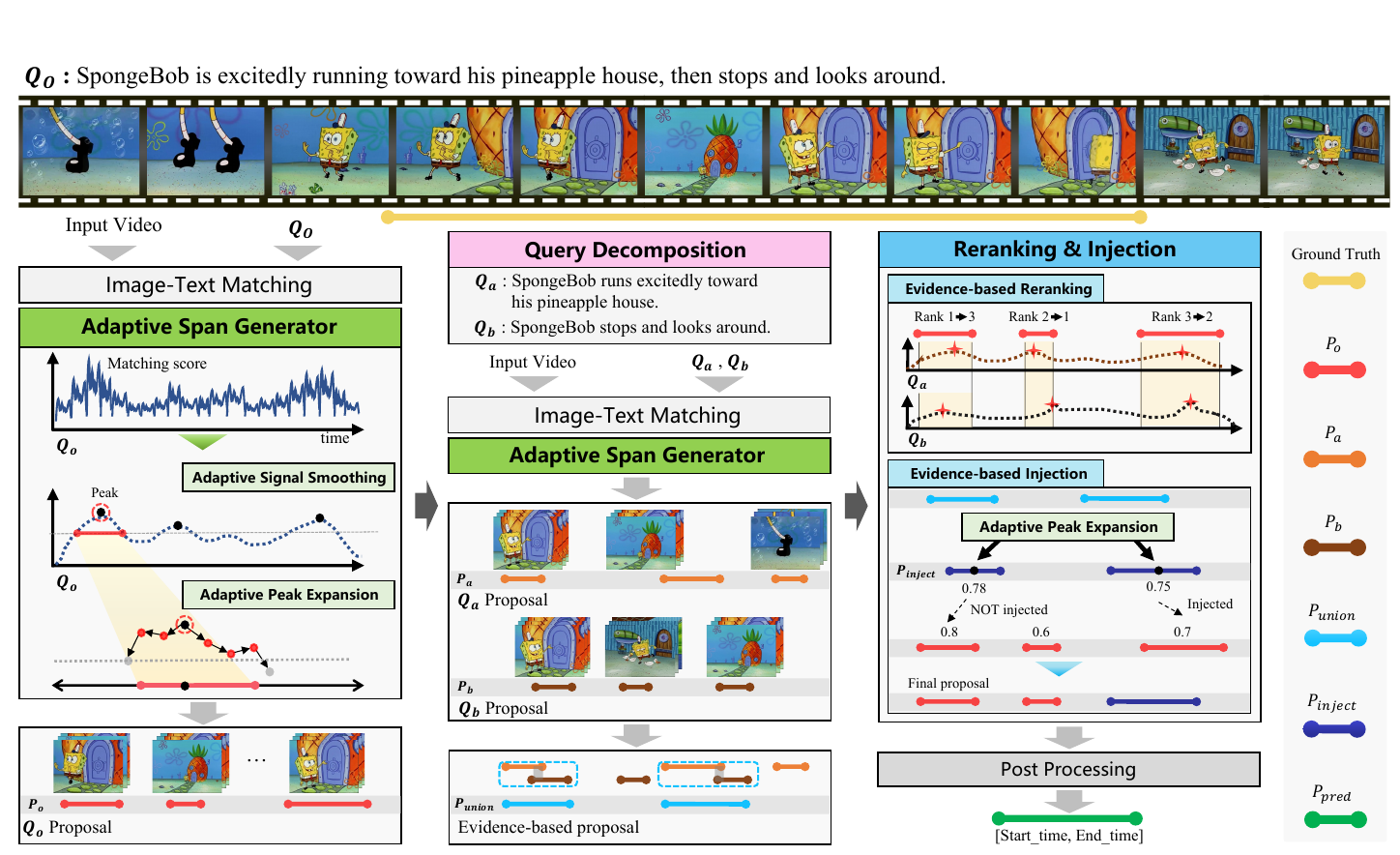}}
\caption{
Overview of the P2S framework. 
First, Adaptive Span Generator (Sec 3.1) adapts to the statistical characteristics of the signal to simultaneously overcome the \textit{search} limitation (temporal search space saturation), thereby detecting robust initial boundaries. 
%
Second, Query Decomposition (Sec 3.2) generates contextual evidence channels that are crucial for addressing the \textit{refine} limitation.
Finally, Evidence-based Proposal Refinement (Sec 3.3) utilizes these channels to precisely rectify `semantic discrepancy' (point vs. span) through reranking and active injection of overlooked intervals.
} 
\label{fig: framework}
\end{figure*}

\section{Related works}
\label{sec:Related works}
\subsection{Supervised Long Video Moment Retrieval}
Supervised LVMR aims to directly learn the relationship between queries and video segments through large-scale annotated datasets such as MAD~\cite{MAD}. 
To this end, sophisticated approaches have been developed to push the boundaries of performance. 
For instance, SnAG~\cite{SnAG} significantly improves retrieval efficiency with a late-fusion design, while ReVisionLLM~\cite{RevisionLLM} attempts to enhance accuracy by mimicking human search processes with a recursive, LLM-driven strategy.
However, despite these advances, even recent models~\cite{SnAG, RevisionLLM, RGNet,vtimellm, TimeChat, LITA,M-Guide,lin2023univtg} face fundamental limitations rooted in the supervised learning paradigm itself.
First, the immense cost of data construction and second, the risk of overfitting to the training domain severely hinder their practicality and scalability. 
Lastly, the absolute performance of these methods remains unsatisfactory due to the inherent complexity of the LVMR task. 
These persistent challenges necessitate a training-free approach, a role our P2S framework is designed to fill by fundamentally addressing the high costs and poor generalization of supervised methods.

\subsection{Zero-Shot Video Moment Retrieval}
Zero-shot VMR offers advantages in scalability and generalize ability by utilizing pre-trained VLMs without additional training. 
However, a major challenge in this paradigm is the \textbf{semantic discrepancy} between concise queries and the visual diversity of actual video content. 
Many recent studies attempt to mitigate this with costly multi-stage pipelines, leveraging LLMs to refine queries~\cite{Moment-GPT, groundinggpt}, generating captions by VLMs~\cite{qwen2.5-vl,video-llama, vid2seq, videollm, video-chatgpt,chatvideo} of candidate segments, or generating and merging complex candidate segments~\cite{zvmr-Gong, vtimellm}.
%
Despite their success on short videos, reliance on costly multi-stage pipelines that involve sequential large-model usage or complex candidate merging~\cite{Moment-GPT, zvmr-Gong, vtimellm, Prompt-based-zvmr} proves impractical for long videos, as they are fail to handle \textbf{temporal search space saturation}.
Our P2S framework directly overcomes these limitations, addressing the \textit{search} challenge (temporal search space saturation) by dynamically adapting to video statistics, and the \textit{refine} challenge (semantic discrepancy) by using decomposed query sub-actions as evidence for refinement.




%% file: sec/3_method.tex
\section{Method}
\begin{figure*}[h!]
\centerline{\includegraphics[width=1\textwidth]{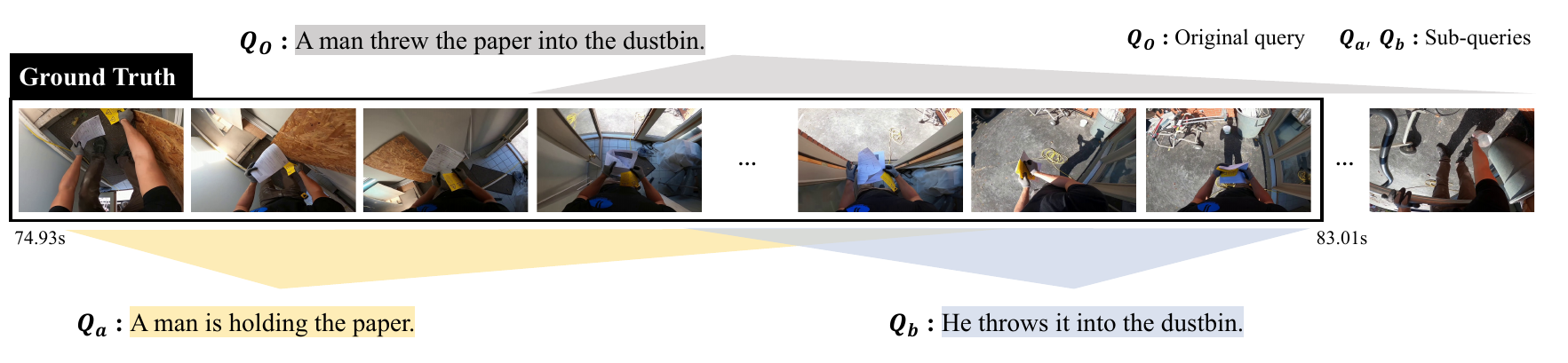}}
\caption{An illustrative example showing the role of sub-queries in Query Decomposition.
The sub-queries $Q_a$ and $Q_b$ act as distinct evidence channels, which successfully localize the start and end boundaries of the full event $Q_o$.
}
\label{fig: QDexamples}
\end{figure*}
In this work, we propose the Point-to-Span (P2S) (illustrated in Figure~\ref{fig: framework}), a novel zero-shot approach designed to efficiently retrieve temporal segments in hour-long videos. 
It is designed to overcome key limitations of existing zero-shot methods in long videos: (1) \textbf{temporal search space saturation} (\textit{search} limitation) and (2) \textbf{semantic discrepancy} (\textit{refine} limitation) challenges during candidate verification. 
These challenges include signal ambiguity, which refers to the issue where a single similarity peak fails to represent the full temporal extent of an event, and the resulting semantic misalignment between the peak signal and the full span semantics.
%
%
To simultaneously address these challenges, we introduce three core modules. 
First, Adaptive Span Generator (Sec.3.1) leverages statistical signal characteristics to identify robust initial boundaries, mitigating the search space saturation. 
%
Second, Contextual Evidence via Query Decomposition (Sec.3.2) decomposes queries to provide richer semantic guidance. 
Finally, Evidence-based Proposal Refinement (Sec.3.3) utilizes this decomposed evidence to precisely refine candidates, resolving the semantic discrepancy between peak signals and full span semantics.

\subsection{Adaptive Span Generator}
The \textit{search} limitation in existing ZS methods often stems from inefficient sliding window approaches and simple fixed thresholding~\cite{zvmr_diwan, CONE}. 
To resolve this inefficiency and address \textbf{temporal search space saturation}, we propose the Adaptive Span Generator (ASG) module.
%
ASG addresses these issues by initiating span expansion only from reliable peaks in a statistically refined signal, consisting of two stages: (1) Adaptive Signal Smoothing and (2) Adaptive Peak Expansion.

\noindent \textbf{Adaptive Signal Smoothing}
To effectively handle diverse noise patterns in long videos, we dynamically adjust the smoothing intensity according to the variability, or `peakiness', of the initial similarity sequence $S_o$. We first calculate the standard deviation of the signal, $\sigma_S = \text{std}(S_o)$. From this, we derive an adaptive ratio, $\tau_r$, which converts the signal's variability into a stabilized scale value used for both smoothing and expansion:
\begin{equation}
\tau_r = 0.5 + 0.5 \cdot \text{sigmoid}(\sigma_S)
\end{equation}
Here, the sigmoid function maps the signal's variability to a stable scale, and the 0.5 offset and scaling ensure $\tau_r$ falls within a practical range, providing a robust base for adaptation. The smoothing window size $w$ is then determined by scaling this ratio with the video's frames-per-second ($\text{fps}$): $w = \text{fps} \cdot \tau_r$. A moving average filter with this calculated window size $w$ is applied to obtain the refined sequence $\tilde{S}$. This induces stronger smoothing for sharper peaks and milder smoothing for flatter ones, ensuring robustness to noise while preserving key semantic information.

\noindent \textbf{Adaptive Peak Expansion}
We identify major peaks $p$ (points) that satisfy temporal distance and prominence constraints from the refined signal $\tilde{S}$ using a standard peak detection algorithm. Conventional methods, which use a fixed threshold for span expansion, are vulnerable in temporal boundary detection. To address this, P2S sets an adaptive expansion threshold, $\tau_{expand}$, by leveraging the previously defined adaptive ratio $\tau_r$. The threshold is determined by considering both the height of each peak $\tilde{S}(p)$ and the global signal variability encapsulated in $\tau_r$:
\begin{equation}
\tau_{expand} = \tilde{S}(p) \cdot \tau_r
\end{equation}
Starting from peak $p$, we expand the span forward and backward to the interval $[t_{\text{start}}^p, t_{\text{end}}^p]$ where the signal $\tilde{S}$ remains higher than $\tau_{expand}$. This expansion enables robust and accurate estimation of temporal boundaries for various types of similarity signals by adaptively adjusting the expansion criteria. This adaptive approach, which simultaneously considers both peak importance and global signal variability, is a core mechanism that enhances the temporal boundary detection accuracy of the P2S framework. We define this expanded interval as a candidate $\text{cand}$ and assign the average score of the refined signal $\tilde{S}$ within this candidate as its base score $s_{\text{base}}(\text{cand})$. We denote this list of candidates as $P_o = \{\text{cand}_1, \text{cand}_2, \dots\}$, representing the candidate group based on the Original Query.

\subsection{Contextual Evidence via Query Decomposition}
While the ASG module in Sec 3.1 efficiently detects robust initial candidate boundaries, it does not fundamentally resolve the second \textit{refine} limitation, the \textbf{semantic discrepancy} problem. 
That is, the initial span found by ASG may be the core moment of the query, but it might not align with the full span of the event required by the query.
For example, a complex action query(``A man opens the package.") implies a span that includes both the `state of holding the package' (start) and the `act of opening' (process), but a single similarity map $S_o$ often highlights only the peak of one of these (e.g., `act of opening'), causing ambiguity.
To resolve this misalignment and obtain rich information for candidate refinement, P2S utilizes an LLM~\cite{qwen2.5} to decompose the original query $Q_o$ into two sub-queries with inherent temporal order: $Q_{a}$ (start-state) and $Q_{b}$ (end-state)\footnote{To ensure semantic consistency and prevent arbitrary decomposition, the LLM is guided by examples and strict guidelines via structured instructions. The detailed strategy and examples are described in Appendix.B}.
As illustrated in Figure~\ref{fig: QDexamples}, we guide the LLM to perform this decomposition using structured instructions. 
These instructions strictly prohibit the generation of unnecessary information and enforce a separation based solely on temporal dependency (e.g., $Q_o$: ``A man threw the paper into the dustbin." → $Q_{a}$: ``A man is holding the paper.", $Q_{b}$: ``He throws it into the dustbin.")
We match the sub-queries ($Q_a$ and $Q_b$) obtained through this decomposition process with the input video to generate respective similarity sequences ($S_a$ and $S_b$).
We term these as independent `evidence channels' that complement the original $S_o$, and utilize them as key clues for refining candidates in the next step.

\subsection{Evidence-based Proposal Refinement}
In the final step, we utilize evidence channels $S_{a}$ and $S_{b}$ to refine the initial candidate group $P_o$ via two complementary mechanisms: Evidence-based Reranking, which adjusts rankings using decisive local evidence, and Evidence-Union-based Injection, which actively re-searches for missed intervals.

\begin{table*}[h!]
\centering
\caption{Main Results on the MAD and MomentSeeker Datasets. P2S demonstrates state-of-the-art performance across both datasets.}
\label{tab:combined_results}
\renewcommand{\arraystretch}{1}
\resizebox{\textwidth}{!}{
\begin{tabular}{l | c | ccccccc | ccccccc} 
\toprule
& \multicolumn{8}{c}{MAD~\cite{MAD}} & \multicolumn{7}{c}{MomentSeeker~\cite{MomentSeeker}} \\
\cmidrule(lr){2-9} \cmidrule(lr){10-16}
\textbf{Model} & \textbf{Training} & \textbf{R1@.1} & \textbf{R5@.1} & \textbf{R1@.3} & \textbf{R5@.3} & \textbf{R1@.5} & \textbf{R5@.5} & \textbf{Avg.}$\uparrow$ & \textbf{R1@.1} & \textbf{R5@.1} & \textbf{R1@.3} & \textbf{R5@.3} & \textbf{R1@.5} & \textbf{R5@.5} & \textbf{Avg.}$\uparrow$ \\
\midrule
M-Guide~\cite{M-Guide} & \checkmark & 9.3 & 18.9 & 4.6 & 13.1 & 2.2 & 7.4 & 9.3 & - & - & - & - & - & - & - \\
VLG-Net~\cite{VLG-NET} & \checkmark & 3.6 & 11.6 & 2.8 & 9.3 & 1.7 & 6.0 & 5.8 &  - & - & - & - & - & - & - \\
M-DETR~\cite{Moment-DETR} & \checkmark & 3.6 & 13.0 & 2.8 & 9.9 & 1.7 & 5.6 & 6.1 & 13.2 & 30.2 &  \underline{9.3} &  \underline{23.8} &  \underline{5.7} &  \underline{13.4} &  \underline{16.5} \\
CONE~\cite{CONE} & \checkmark & 8.9 & 20.5 & 6.9 & 16.1 & 4.1 & 9.6 & 11.0 & - & - & - & - & - & - & -  \\
SnAG~\cite{SnAG} & \checkmark & 10.3 & 24.4 & 8.5 & \textbf{20.6} & 5.5 & \textbf{13.7} & 13.8 & - & - & - & - & - & - & -  \\
RGNet~\cite{RGNet} & \checkmark & 12.4 & 25.1 & \underline{9.5} & 18.7 & 5.6 & 10.9 & 13.7 & - & - & - & - & - & - & - \\
VTimeLLM~\cite{vtimellm} & \checkmark & 1.4 & 3.1 & 1.3 & 2.5 & 0.6 & 1.1 & 1.7 & - & - & - & - & - & - & - \\
RevisionLLM~\cite{RevisionLLM} & \checkmark & \textbf{15.0} & \underline{25.1} & \textbf{11.0} & \underline{18.8} & \textbf{5.8} & 10.5 & \underline{14.4} & - & - & - & - & - & - & - \\
\midrule
CLIP~\cite{CLIP} & - & 6.6 & 15.1 & 3.1 & 9.9 & 1.5 & 5.4 & 6.9 &  \underline{14.7} &  \underline{39.0} & 3.3 & 20.6 & 1.2 & 10.2 & 14.8 \\
\midrule
\textbf{Ours} & - & \underline{13.5} & \textbf{28.8} & 9.3 & 18.6 & \textbf{5.8} & \underline{11.1} & \textbf{14.5} & \textbf{49.3} & \textbf{66.8} & \textbf{30.8} & \textbf{44.7} & \textbf{19.5} & \textbf{26.6} & \textbf{39.3} \\
\bottomrule
\end{tabular}}
\end{table*}

\noindent \textbf{Evidence-based Reranking}
We re-adjust the ranking by quantifying how well the initial candidate $cand \in P_o$ matches each element ($Q_{a}, Q_{b}$) of the decomposed query. 
First, we calculate the evidence bonus score $s_{bonus}(cand)$ by summing the maximum peaks of the unsmoothed $S_a$ and $S_b$ signals within each candidate's interval:
\begin{equation}
    s_{bonus}(cand) = \max(S_{a}^{cand}) + \max(S_{b}^{cand})
\end{equation}
Using maxima specifically captures short but decisive evidence signals (specific states or actions) that might be diluted in averages.
The final score is obtained by adding a weighted bonus to the base score (We empirically set $\beta$ = 0.5 in our experiments, as detailed in the Appendix.A).
\begin{equation}
  s_{rerank}(cand) = s_{base}(cand) + \beta \cdot s_{bonus}(cand)
\end{equation}
This bonus increases the score as the candidate interval strongly contains evidence for both the preceding context ($S_{a}$) and the core action ($S_{b}$) of the event, thereby raising the rank of intervals that more faithfully reflect the entire event composition. 
We define this calibrated list as $P'_{o}$.

\noindent \textbf{Evidence-Union-based Injection}
Reranking only adjusts the order of existing candidates and cannot recover important intervals missed in the initial stage. 
In particular, intervals that were ambiguous in $S_o$ but appear strongly in common in both $S_a$ and $S_b$ are highly likely to be actual important events. 
We introduce an injection mechanism to actively discover these `missed candidates'.

First, we independently apply ASG to $S_{a}$ and $S_{b}$ to generate candidate groups $P_{a}$ and $P_{b}$.
We then identify the temporal union of overlapping pairs from $P_{a}$ and $P_{b}$ as the `high-confidence region' $R_{\cup}$.
This region pinpoints potential event intervals with increased reliability through cross-verification.
Next, we re-applies the ASG module within this `high-confidence region ($R_{\cup}$)' while maintaining the global standard deviation ($\sigma_{S}$). 
This allows for the discovery of new candidate groups $P_{inject}$ that are locally prominent within $R_{\cup}$ while maintaining the statistical consistency of the entire video.
Re-discovered candidates $cand \in P_{inject}$ are assigned a final score $s_{rerank}(cand)$ by summing $s_{base}(cand)$ and $s_{bonus}(cand)$.
Finally, we integrate this $P_{inject}$ with the previous reranking result $P'_{o}$ and apply final NMS to select the most reliable final candidate list $P_{final}$, complementing the limitations of the initial search.
This Rerank-and-Inject mechanism is a key contribution that refines candidates self-sufficiently within the pipeline and enhances completeness by dually utilizing temporal evidence decomposed by LLM~\cite{qwen2.5} for both ranking bonuses and new candidate search guides.

%% file: sec/4_experiments.tex
\section{Experiments}
\label{sec:Experiments}

\subsection{Experimental Settings and Details} 
\noindent \textbf {Dataset}
In this study, we use two challenging long-form video moment retrieval (LVMR) benchmarks, MAD~\cite{MAD} and MomentSeeker~\cite{MomentSeeker}, to validate the performance of the P2S framework.
\noindent \textbf{MAD} is a large-scale benchmark based on approximately 1,200 hours of full-length movies. The average video duration is very long at 110.8 minutes, whereas the average length of target moments is only 4.1 seconds. This extremely low moment-to-video ratio is a core challenge that makes accurate temporal grounding difficult. 
\textbf{MomentSeeker} is a benchmark for retrieving moments based on various tasks (causal reasoning, action recognition, etc.) from long videos averaging over 20 minutes. Among these, we select and use the text-only query dataset that falls within the scope of this study.

\noindent \textbf{Evaluation Metrics.} 
We measure whether any of the top $N \in \{1, 5\}$ predictions match the ground truth moment using a Temporal Intersection over Union (tIoU) threshold $K \in \{0.1, 0.3, 0.5\}$. This standard metric is denoted as R@N, tIoU=$K$. Furthermore, we use Avg., the average of all R@N, tIoU=$K$ metrics, as the primary performance indicator to compare the overall performance of the models.



\noindent \textbf{Implementation Details}
We utilized the standard pre-trained CLIP ViT-L/14~\cite{CLIP} for both visual and text feature extraction without fine-tuning. Videos were downsampled to 5 FPS, and we used the token feature from the visual encoder. For Query Decomposition, we employed Qwen2.5-3B~\cite{qwen2.5}. In the Adaptive Span Generator module, we used scipy.signal.find\_peaks for peak detection. We set the minimum temporal distance to 1 second to mitigate high-frequency noise and applied an empirical prominence threshold of 0.05 to filter out insignificant local maxima. The balancing weight $\beta$ for Evidence-based Reranking (Eq. 4) was set to 0.5. Finally, we applied Non-Maximum Suppression (NMS) with a tIoU threshold of 0.8 to remove redundant proposals. A sensitivity analysis on these key hyperparameters is provided in the Appendix.A.

\subsection{Main Results} 

Table \ref{tab:combined_results} presents a performance comparison of the proposed P2S model against existing SOTA models on the MAD~\cite{MAD} and MomentSeeker~\cite{MomentSeeker} benchmarks. 

\noindent \textbf{Results on MAD.} 
On the MAD dataset, P2S recorded an Avg. score of 14.5\%. This slightly outperforms the 14.4\% of RevisionLLM~\cite{RevisionLLM}, the latest supervised learning model that utilizes massive training resources. Achieving performance comparable to or better than SOTA supervised models without any training demonstrates the efficiency and power of the P2S framework. 
Analyzing detailed metrics, P2S shows slightly lower R1@K performance compared to RevisionLLM, but higher R5@K performance. This suggests a key operational difference: supervised models are optimized to pinpoint a single ground-truth, reflecting their training bias. In contrast, our zero-shot P2S excels at capturing not just a single primary peak, but also other semantically relevant signals to identify multiple plausible candidates. This robust candidate generation is the key factor driving its superior R5@K and overall average (Avg.) performance.
%
\begin{figure}[htb!]
\centerline{\includegraphics[width=1\columnwidth]{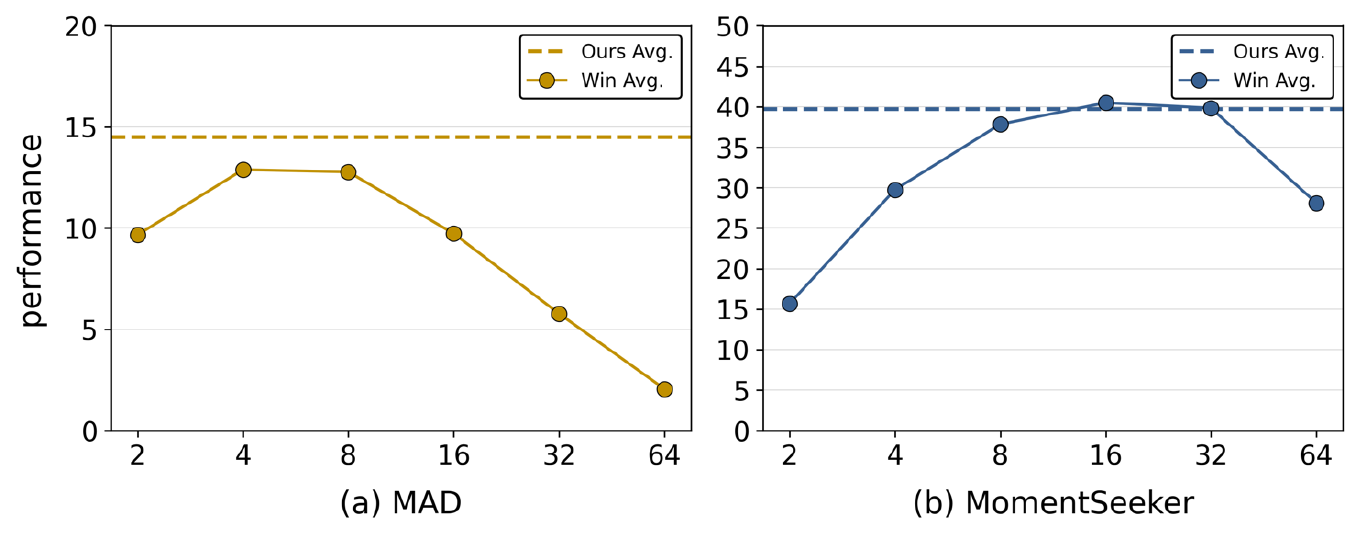}}
\caption{
Performance (Avg. R1, R5) comparison between the fixed window method (Win Avg.) and our ASG (Ours Avg.). 
Our adaptive method demonstrates robust performance across datasets, whereas the fixed window method is highly sensitive to window size and requires costly, per-dataset tuning. 
} 
\label{tab:ablation_step1}
\end{figure}
\noindent \textbf{Results on MomentSeeker.} 
On the MomentSeeker dataset, P2S achieved an Avg. of 39.3\%, significantly outperforming both the baseline CLIP~\cite{CLIP} (14.8\%) and the supervised model M-DETR~\cite{Moment-DETR} (16.5\%). 
As MomentSeeker is a recent benchmark, performance from most recent supervised models~\cite{RevisionLLM, SnAG, SOONet, RGNet} is not yet available. 
Nevertheless, this dataset exemplifies a scenario where supervised methods are inherently challenged: it offers a limited amount of training data. 
This highlights the critical dependency of supervised learning on massive datasets, a practical bottleneck that P2S avoids. 
In contrast, our zero-shot P2S thrives in such data-scarce environments, proving its superior scalability and real-world viability.
\subsection{Ablation Study}
We conducted a three-step ablation study to verify how each component of the proposed \textbf{P2S} framework contributes to the final performance. This study sequentially analyzes the effectiveness of our \textit{Adaptive Span Generator (ASG)} module, the robustness of its adaptive components, and the contribution of the \textit{Query Decomposition (QD)}-based refinement modules.
\begin{table}[h]
\centering
\caption{
Cross-dataset robustness comparison for two adaptive mechanisms: 
\textbf{Adaptive Signal Smoothing} and \textbf{Adaptive Peak Expansion}. 
Results (Avg. R1, R5) are reported separately on MAD and Charades-STA datasets.}
\label{tab:ablation_adaptive}
\renewcommand{\arraystretch}{1}
\resizebox{\columnwidth}{!}{%
\begin{tabular}{l  l  c  c}
\toprule
\textbf{Model} & \textbf{Tuned on} & \textbf{MAD Avg.} & \textbf{Charades-STA Avg.} \\
\midrule
\multicolumn{4}{l}{\textbf{\textit{Adaptive Signal Smoothing}}} \\
\midrule
Fixed-Params          &  MAD & 14.24  & 54.69 \\
Fixed-Params &  Charades-STA & 12.96  & \textbf{60.55}\\
\midrule
\multicolumn{4}{l}{\textbf{\textit{Adaptive Peak Expansion}}} \\
\midrule
Fixed-Params          &  MAD & \underline{14.39}  & 55.71 \\
Fixed-Params &  Charades-STA & 12.99  & 59.35 \\
\midrule
\textbf{Adaptive (ASG)} & -- & \textbf{14.51}  & \underline{59.41} \\
\bottomrule
\end{tabular}
}
\end{table}
\noindent \textbf{Effectiveness of Adaptive Span Generator.}
First, we verified whether the ASG in Sec 3.1 is more effective than conventional fixed approaches. 
Figure \ref{tab:ablation_step1} shows the comparison results of the fixed sliding window method with various sizes and the ASG method on the MAD and MomentSeeker datasets. 
The fixed window method is highly sensitive to window size, showing different optimal sizes and significant performance variations for each dataset.
As shown in Figure \ref{tab:ablation_step1}, size 4 is optimal for (a) MAD, but 16 is optimal for (b) MomentSeeker.
Even though the fixed window method showed slightly higher performance than ASG at window size 16 for MomentSeeker, finding this requires search costs for all window sizes. 
In contrast, our ASG achieved consistent and robust performance comparable to or exceeding the optimal fixed window performance on both datasets with a single setting, without such expensive hyperparameter tuning processes. 
This proves the superior ability of ASG to dynamically determine optimal intervals according to diverse signal characteristics.


\noindent \textbf{Robustness of Adaptive Mechanisms.}
We further analyze the robustness of the two adaptive components - \textit{Adaptive Signal Smoothing} and \textit{Adaptive Peak Expansion}.
Because video lengths and signal dynamics vary drastically across datasets, a fixed parameter cannot flexibly handle all environments.
To validate this, we conducted cross-dataset evaluations on MAD ($\approx 110$ min) and Charades-STA ($\approx 30$ sec), which exhibit extremely different temporal characteristics.
As shown in Table \ref{tab:ablation_adaptive}, in both modules, fixed-parameter models with hyperparameters tuned on a specific dataset show a clear performance collapse when applied to another dataset with different characteristics.
For example, in the Adaptive Signal Smoothing module, the parameter tuned on MAD achieves 14.24\% on MAD but drops to 12.96\% on Charades-STA, while the parameter tuned on Charades-STA scores 60.55\% on Charades-STA but only 54.69\% on MAD.
In contrast, the adaptive variant maintains stable performance around 14.51\% and 59.41\% across both datasets without any additional tuning.
These results confirm that adaptive parameterization is not merely beneficial but essential for achieving robust zero-shot performance across diverse real-world scenarios.
\begin{table}[h]
\centering
\caption{Contribution analysis of Query Decomposition (QD)-based refinement modules on MomentSeeker. For (a), (b) each ER, EI denotes Evidence-based Reranking and Injection.}
\label{tab:ablation_components}
\renewcommand{\arraystretch}{1.1}
\resizebox{1\columnwidth}{!}{%
\begin{tabular}{lccccc}
\toprule
\textbf{Model} & \textbf{R1@0.1} & \textbf{R1@0.3} & \textbf{R5@0.1} & \textbf{R5@0.3} & \textbf{Avg.} \\
\midrule
ASG & 45.92 & 28.23 & 64.21 & 40.95 & 36.51 \\
(a)  (+ER) & \textbf{48.31} & \underline{30.42} & \underline{66.40} & 41.95 & \underline{38.10} \\
(b)  (+EI) & 42.35 & 28.23 & 66.21 & \underline{44.14} & 37.38 \\
\textbf{(c) P2S(+ER\&EI)} & \underline{47.32} & \textbf{30.82} & \textbf{66.80} & \textbf{44.73} & \textbf{39.30}\\
\bottomrule
\end{tabular}}
\end{table}
\noindent \textbf{Effectiveness of Query Decomposition Refinement.}
Finally, we analyzed how effectively the proposed Query Decomposition(QD)-based refinement modules enhance the strong baseline performance of \textit{Adaptive Span Generator (ASG)} on the MomentSeeker dataset.
Starting from ASG, we sequentially added two components: (a) \textit{Evidence-based Reranking(ER)}, (b) \textit{Evidence-based Injection(EI)}, and (c) both combined, forming our final \textbf{P2S} framework.
Table \ref{tab:ablation_components} shows that the ASG baseline already achieved a solid average performance of 36.51\%.
Adding ER notably improved precision by refining the ranking of existing candidates, while EI contributed mainly to recall by recovering moments missed in the initial search.
When both modules were integrated, the resulting \textbf{P2S} framework achieved the highest overall performance, confirming the complementary effects of ER and EI for robust and balanced moment localization.
Finally, (c) the combined model, \textbf{P2S (+ER\&EI)}, achieved the highest overall average of 39.30\%, demonstrating the complementary nature of the two modules - ER refines existing candidate ranks, while EI supplements missing ones.
This synergy completes the full P2S architecture for robust and balanced moment localization.

\begin{figure*}[htb!]
\centerline{\includegraphics[width=1.0\textwidth]{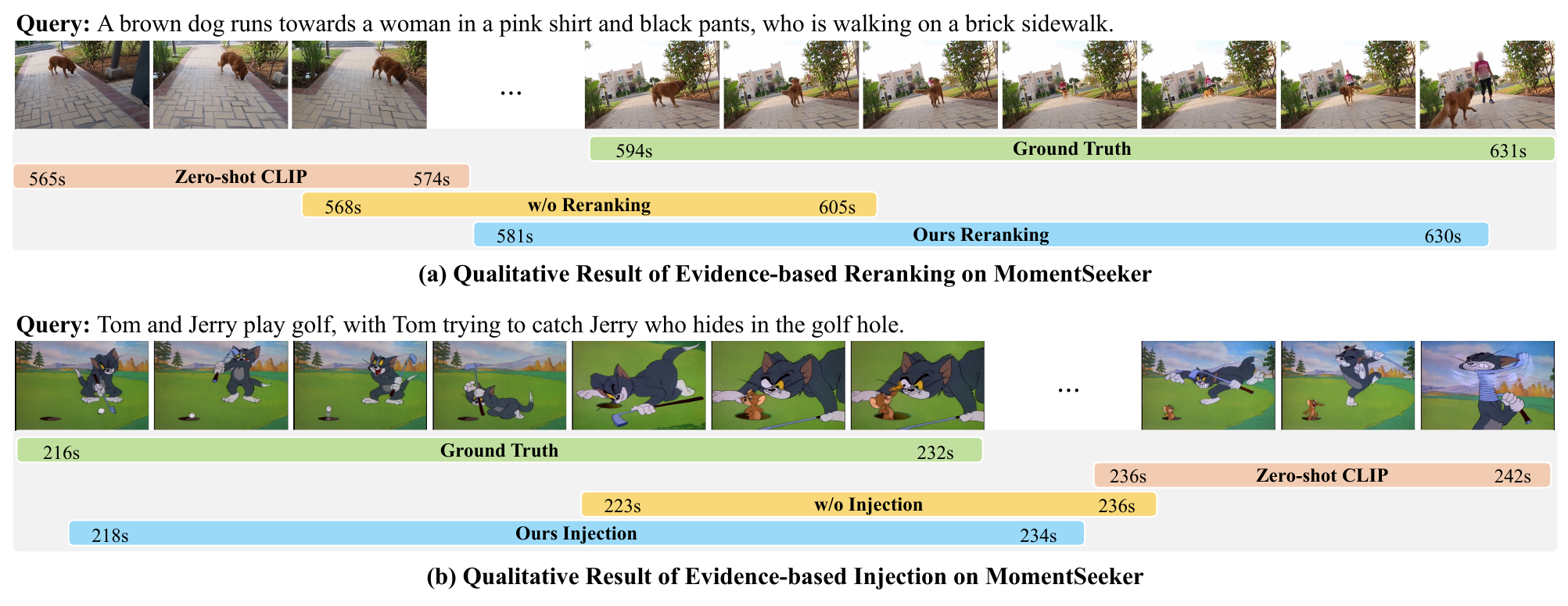}}
\caption{Qualitative result analysis demonstrating the final results of our model while showing the comparison with Zero-shot CLIP baseline and the effect of (a) reranking and (b) injection.
} 
\label{fig:qr}
\end{figure*}
\begin{table}[t]
    \centering
    \caption{Comparison of query decomposition strategies on the MAD dataset. Our LLM-based semantic approach is compared against naive word-count and rule-based splitting baselines. (Avg. denotes the average of  R1, R5) }
    \label{tab:ablation_decomposition}
    \resizebox{1.0\linewidth}{!}{%
        \begin{tabular}{l|ccccc}
            \toprule
            \textbf{Decomposition Strategy} & \textbf{R1@0.1} & \textbf{R1@0.3} & \textbf{R5@0.1} & \textbf{R5@0.3} & \textbf{Avg.} \\
            \midrule
            Naive-Split (Word count) & 9.1 & 20.2 & 6.5  & 12.7 & 10.2 \\
            Rule-based (Delimiters) & 11.7 & 24.5 & 7.6  & 15.4 & 13.1 \\
            \textbf{Ours (LLM-Semantic)} & \textbf{13.5} & \textbf{28.8} & \textbf{9.3} & \textbf{18.6} & \textbf{14.5} \\
            \bottomrule
        \end{tabular}%
    }
\end{table}
\subsection{Further Analysis}
We conduct an in-depth analysis to validate that our Query Decomposition is not merely a heuristic, but a robust mechanism essential for addressing the semantic discrepancy in long videos.

\noindent \textbf{Efficacy of Semantic-aware Decomposition.} 
We investigate the impact of how the query is decomposed by comparing our LLM-based semantic decomposition with two baselines: (i) Naive-Split, which simply bisects the query based on word count, and (ii) Rule-based, which splits queries based on explicit delimiters (e.g., `and’, `while’). 
As presented in Table~\ref{tab:ablation_decomposition}, Naive-Split performs poorly, achieving an Avg. score of 10.2\%, by breaking semantic integrity (e.g., splitting ``turns around” into ``turns” and ``around”), thereby generating noisy guiding signals. 
Rule-based splitting offers improvement with an Avg. score of 13.1\%, but fails when explicit delimiters are absent or ambiguous.
In contrast, our LLM-based approach consistently outperforms both baselines, achieving the best performance with an Avg. score of 14.5\% by preserving complete semantic units (e.g., Subject-Verb-Object structures) for each sub-event. 
This ensures that every sub-query acts as a reliable anchor for precise span refinement regardless of query complexity.

\noindent \textbf{Analysis of Throughput and Accuracy Trade-off.} We analyzed the inference throughput of the P2S framework in relation to the query decomposition method. 
When using simple decomposition methods that do not involve LLM calls, such as `Naive-Split' or `Rule-based', the framework achieves a high throughput, processing approximately 40 videos per second (based on an NVIDIA A6000 GPU). 
However, as clearly confirmed in the previous analysis (Table~\ref{tab:ablation_decomposition}), both of these methods fail to guarantee the semantic integrity of the query, leading to significant performance degradation. 
In contrast, applying our LLM-based semantic decomposition incurs a fixed, one-time setup cost of approximately 0.5 seconds per query, reducing the effective throughput to 2 videos per second. 
This 0.5-second overhead is a `fixed entry cost' that is essentially required to prevent catastrophic performance degradation and secure semantic robustness. 
Considering the complexity of the ZLVMR task, we conclude that this cost is a highly reasonable and necessary trade-off to pay for accuracy.

\subsection{Qualitative Results}
To further illustrate the effect of each refinement component, we present qualitative examples in Figure~\ref{fig:qr}, which compare our P2S framework with zero-shot CLIP and its non-refined variants.  
In (a), the example highlights the effect of the \textit{Evidence-based Reranking (ER)} module.  
While the zero-shot CLIP baseline roughly identifies the relevant region but fails to align temporal boundaries, and the version without reranking remains imprecise, our reranked prediction accurately matches the ground truth with a high tIoU.  
In (b), we demonstrate the impact of the \textit{Evidence-based Injection (EI)} module.  
Here, the zero-shot baseline completely misses the relevant moment, and the version without injection fails to recover it, whereas our full method successfully detects the correct segment by injecting additional candidate intervals derived from complementary query evidence.  
Together, these examples show that \textit{ER} refines boundary precision while \textit{EI} enhances recall, confirming that the two modules contribute complementary improvements for robust temporal localization.


%% file: sec/5_conclusion.tex
\section{Conclusion}
\label{sec:Conclusion}

This paper has identified the cost and generalization limitations of existing supervised learning approaches in hour-long video environments and has also revealed that the Zero-Shot paradigm, the only viable alternative, faces fundamental challenges in the \textit{search} and \textit{refine} phases. 
This study proposes the P2S framework, which directly addresses these core challenges in a training-free manner. 
Remarkably, P2S achieved performance surpassing that of supervised SOTA models trained with massive data, empirically demonstrating that efficient, robust, and scalable zero-shot LVMR is practically feasible.

\section{Limitations}

While P2S has demonstrated high potential, it has two major limitations. First, its performance is fundamentally upper-bounded by the quality of the pre-trained features (e.g., CLIP); detection can fail if all evidence channels ($S_o, S_a, S_b$) produce weak signals. Second, Query Decomposition increases inference overhead (Sec 4.4), creating a trade-off between accuracy and efficiency. Thus, future work should address these by exploring better long-video features and developing lightweight, LLM-free refinement mechanisms to optimize this trade-off.


%% file: sec/X_suppl.tex
\appendix

\begin{appendix}


\section{Hyperparameter Sensitivity Analysis}
\label{sec:appendix_sensitivity}

In this section, we provide a detailed analysis of the robustness of our proposed framework against variations in key hyperparameters on the MomentSeeker dataset. We demonstrate that the model maintains high performance across broad ranges of parameter settings, confirming its reliability and generalizability.

\noindent \textbf{Robustness to Peak Prominence.}
The peak prominence parameter ($pm$) is crucial for filtering noise and local maxima in the similarity curve. 
$\text{Table \ref{tab:prominence_sensitivity}}$ demonstrates that the model exhibits low sensitivity to the peak prominence threshold ($pm$). The performance remains consistent across all tested values, indicating that our framework is robust to variations in peak filtering criteria. We select $pm=0.05$ as the default setting, as it yields optimal performance among the stable configurations.

\noindent \textbf{Robustness to Minimum Temporal Distance.}
The minimum temporal distance ($mtd$) parameter serves to filter out high-frequency noise by constraining the peak separation. As shown in Table~\ref{tab:distance_sensitivity}, the performance is highly robust across small $mtd$ values (e.g., $\le 2.0$), indicating that the model effectively handles noise without requiring aggressive filtering. However, excessively large values (e.g., 5.0) lead to a performance drop by suppressing valid peaks in proximity. Therefore, we select $mtd=1.0$ as a standard setting that ensures stability.

\noindent \textbf{Robustness to Reranking Weight.}
The parameter $\beta$ (in Eq. 4) governs the influence of the evidence bonus score derived from decomposed sub-queries during the refinement phase. As shown in $\text{Table \ref{tab:beta_sensitivity_final}}$, the model exhibits remarkable robustness in the range $\beta \in [0.3, 0.9]$, where the average performance fluctuates by less than 0.3 points. This demonstrates that the score fusion is highly tolerant of the specific weight assigned to the evidence. Performance is maximized at $\beta=0.5$ and $\beta=0.7$; we select $\beta=0.5$ as the central stable point yielding the highest overall performance.

\begin{table}[h]
    \centering
    \caption{Robustness analysis of the peak prominence threshold ($pm$). The model exhibits low sensitivity to $pm$, showing consistent performance across all tested values.}    
    \label{tab:prominence_sensitivity}
    \resizebox{\columnwidth}{!}{%
    \begin{tabular}{lccccc}
        \toprule
        $pm$ & R1@0.1 & R1@0.3 & R5@0.1 & R5@0.3 & Avg. \\
        \midrule
        0.03 & 49.11 & 30.82 & 66.60 & 42.15 & 38.54 \\
        0.04 & 49.30 & 31.01 & 66.60 & 42.15 & 38.57 \\
        \textbf{0.05} & \textbf{49.30} & \textbf{31.01} & \textbf{67.00} & \textbf{42.15} & \textbf{38.70} \\
        0.06 & 48.71 & 31.01 & 65.01 & 41.55 & 38.01 \\
        0.07 & 48.91 & 31.01 & 64.41 & 41.35 & 37.94 \\
        \bottomrule
    \end{tabular}}
\end{table}
\begin{table}[h]
    \centering
    \caption{Robustness analysis of the minimum temporal distance ($mtd$). The performance is highly stable around $mtd=1.0\text{s}$, which balances noise filtering and moment fragmentation.}    
    \label{tab:distance_sensitivity}
    \resizebox{\columnwidth}{!}{%
    \begin{tabular}{lccccc}
        \toprule
        $mtd$ & R1@0.1 & R1@0.3 & R5@0.1 & R5@0.3 & Avg. \\
        \midrule
        0.5 & 49.30 & 31.01 & 67.00 & 42.15 & 38.70  \\
        \textbf{1.0} & \textbf{49.30} & \textbf{31.01} & \textbf{67.00} & \textbf{42.15} & \textbf{38.70} \\
        2.0 & 49.30 & 31.01 & 67.00 & 42.15 & 38.70 \\
        5.0 & 48.91 & 31.01 & 66.40 & 41.95 & 38.47 \\
        10.0 & 48.51 & 30.62 & 66.40 & 41.35 & 38.07 \\
        \bottomrule
    \end{tabular}}
\end{table}
\begin{table}[h]
    \centering
    \caption{Robustness analysis of the reranking weight $\beta$. The performance remains stable across $\beta \in [0.1, 0.9]$, confirming low sensitivity to the evidence bonus weight.}    
    \label{tab:beta_sensitivity_final}
    \resizebox{\columnwidth}{!}{%
    \begin{tabular}{lccccc}
        \toprule
        $\beta$ & R1@0.1 & R1@0.3 & R5@0.1 & R5@0.3 & Avg. \\
        \midrule
        0.1 & 46.52 & 28.83 & 65.61 & 41.75 & 37.31 \\
        0.3 & 48.11   &  30.22  & 66.80   & \textbf{42.35}   & 38.24   \\
        \textbf{0.5} & \textbf{49.30}   & \textbf{31.01}   & \textbf{67.00}   & 42.15   & \textbf{38.70}   \\
        0.7 & 49.11   & 31.01   & 66.80   & \textbf{42.35}   &  \textbf{38.70}  \\
        0.9 & 48.51   & 30.22   & 66.80   & \textbf{42.35}   & 38.44  \\
        \bottomrule
    \end{tabular}}
\end{table}
\begin{table}[h]
\centering
\caption{Robustness analysis of the Non-Maximum Suppression (NMS) threshold. Performance is highly stable around a threshold of 0.8, effectively balancing the removal of redundant proposals and the preservation of distinct moments.}
\label{tab:beta_sensitivity}
\resizebox{\columnwidth}{!}{%
\begin{tabular}{lccccc}
\toprule
$NMS$ & R1@0.1 & R1@0.3 & R5@0.1 & R5@0.3 & Avg. \\
\midrule
0.7 & 48.79 & 30.82 & 66.51 & 41.95 & 38.54 \\
0.75 & 49.30 & 31.01 & 67.00 & 42.15 & 38.70 \\
\textbf{0.8} & \textbf{49.30} & \textbf{31.01} & \textbf{67.00} & \textbf{42.15 }& \textbf{38.70} \\
0.85 & 49.11 & 30.62 & 66.80 & 41.75 & 38.62 \\
0.9 & 48.51 & 30.62 & 66.40 & 41.95 & 38.44 \\
\bottomrule
\end{tabular}}
\end{table}

\noindent \textbf{Robustness to NMS Threshold.}
The Non-Maximum Suppression (NMS) threshold filters overlapping temporal proposals to reduce redundancy. As shown in Table \ref{tab:beta_sensitivity}, our model is robust to variations in this threshold, achieving peak performance at both $0.75$ and $0.8$. A low threshold can mistakenly suppress valid, close proposals, while a high threshold may fail to remove redundant detections. We select an NMS threshold of $0.8$, which lies in the center of the most stable and high-performing region.

\begin{figure*}[h!]
\centerline{\includegraphics[width=0.85\textwidth]{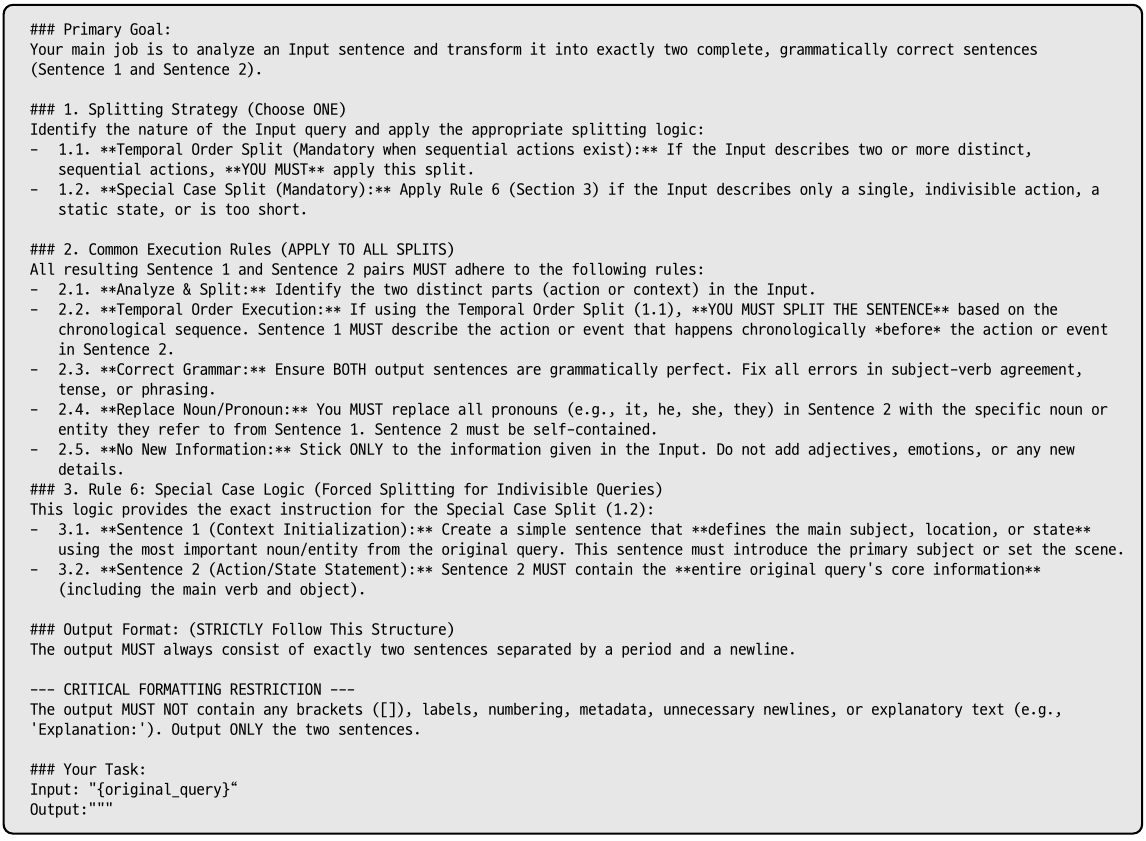}}
\caption{Query Decomposition Prompt.
}
\label{fig: QDexamples}
\end{figure*}

\section{Query Decomposition Details}
\label{sec:appendix_b}
In this section, we elaborate on the prompt engineering constraints for Query Decomposition and provide the rationale behind our binary splitting strategy.

\subsection{Prompt Engineering for Query Decomposition}
\label{sec:appendix_b}
The integrity of the Query Decomposition (QD) module hinges on generating two sub-queries, $Q_a$ and $Q_b$, that strictly adhere to chronological order and semantic robustness. 
To mitigate potential risks of arbitrary decomposition or hallucination, we engineered a structured prompt with explicit hard constraints, as detailed in Figure 1. 
Specifically, we enforce three key safeguards: 
(1) \textbf{Strict Temporal Sequencing} to ensure $Q_a$ chronologically precedes $Q_b$; 
(2) \textbf{Semantic Self-Containment} via mandatory pronoun-to-noun replacement; and 
(3) \textbf{Atomic Query Handling} (Rule 6) to force a context-action split for atomic queries. 
These constraints guarantee that the generated similarity channels ($S_a$ and $S_b$) serve as distinct and reliable temporal evidence.

\subsection{Rationale for Binary Decomposition Strategy} 
While fine-grained decomposition (i.e., $N > 2$) may seem to offer more detailed semantic guidance, it often leads to \textbf{semantic fragmentation}, where sub-queries become too short and ambiguous, introducing noise rather than clarity. Furthermore, it linearly increases the computational overhead.
Our framework overcomes the limitation of binary splitting through the \textbf{Evidence-Union} mechanism. By anchoring the start ($Q_a$) and end ($Q_b$) of an event, the system identifies a high-confidence interval. The \textbf{Adaptive Span Generator (ASG)} then leverages signal continuity to naturally interpolate the temporal gap. Consequently, intermediate actions that are not explicitly split in the text are \textbf{implicitly merged} into the final span due to their temporal adjacency. This confirms that the binary split is the most efficient and robust strategy for handling complex queries.
\section{Computational Efficiency Analysis}
\label{sec:appendix_c}

\begin{figure*}[t]
    \centering
    \includegraphics[width=0.95\textwidth]{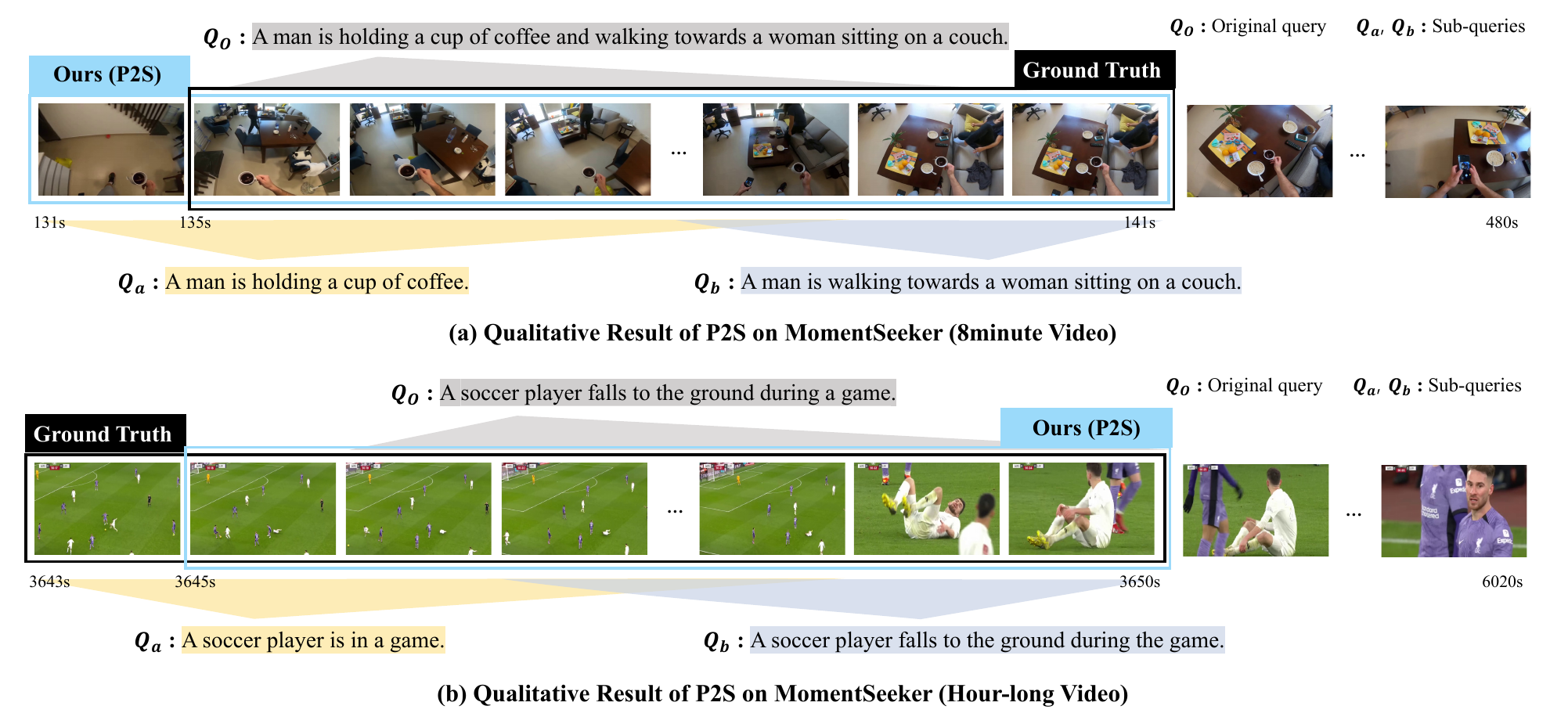}
    \caption{Additional qualitative visualization. (a) Resolving \textbf{semantic discrepancy} in a complex compositional query on an 8-minute video by utilizing decomposed evidence channels ($Q_a, Q_b$) to align sub-events precisely. (b) Overcoming \textbf{temporal search space saturation} in an hour-long video, where the model filters extensive background context ($Q_a$: in a game) to pinpoint the specific short action ($Q_b$: falls).}
    \label{fig:cvpr_fig8}
\end{figure*}

\begin{table}[t]
    \centering
    \caption{Runtime breakdown per query for processing hour-long videos. Complexity is measured with respect to the video length $T$. Note that Query Decomposition relies solely on the query text, maintaining $O(1)$ complexity regardless of $T$.}
    \label{tab:runtime}
    \resizebox{\columnwidth}{!}{%
    \begin{tabular}{lccc}
        \toprule
        \textbf{Component} & \textbf{Complexity} & \textbf{Time (s)} & \textbf{Note} \\
        \midrule
        Query Decomposition & $O(1)$ & $\sim$0.50 & Generation \\
        Adaptive Span Generation & $O(T)$ & $<$ 0.05 & 
        Dot Product \\
        Refinement (Matching) & $O(T)$ & $<$ 0.05 & Dot Product \\
        \midrule
        \textbf{Total Inference} & $O(T)$ & \textbf{$\sim$0.60} & Per video \\
        \bottomrule
    \end{tabular}
    }
\end{table}

\noindent \textbf{Runtime Analysis on P2S.}
We provide a detailed breakdown of the online inference phase to demonstrate the real-time efficiency of P2S on the MAD dataset. Measurements were conducted on a single NVIDIA A6000 GPU, averaged over 1-hour videos (downsampled to 5 FPS).
As shown in Table~\ref{tab:runtime}, the maximum total inference time per video is approximately 0.60 seconds. The primary latency stems from Query Decomposition ($\sim$0.5s), which incurs a fixed generation cost independent of video duration ($O(1)$). In contrast, the subsequent steps, Adaptive Span Generation and Refinement, scale linearly with video length ($O(T)$). However, since these modules rely on highly optimized matrix operations (e.g., dot products) rather than heavy model forward passes, their computational overhead is negligible ($<0.05s$). This demonstrates that P2S is a feasible solution for real-world applications involving long-form videos.

\begin{table}[h]
    \centering
    \caption{Performance comparison on the MAD dataset across various LLM architectures and scales. The results (Avg. of R1 and R5) show minimal variance ($<0.5\%$) despite differences in model size, validating the model-agnostic robustness of P2S.}
    \label{tab:llm_generalization}
    \resizebox{0.8\columnwidth}{!}{%
    \begin{tabular}{lccc}
        \toprule
        \textbf{Model} & \textbf{Params} & \textbf{Avg. (R1, R5)} & \textbf{Latency (s)} \\
        \midrule
        Qwen2.5 & 1.5B & 14.2 & 0.32 \\
        \textbf{Qwen2.5 (Ours)} & \textbf{3B} & \textbf{14.5} & \textbf{0.34} \\
        Qwen2.5 & 7B & 14.5 & 1.04 \\
        \midrule
        Llama-3.2 & 3B & 14.2 & 0.52 \\
        Llama-3.1 & 8B & 14.3 & 0.95 \\
        \midrule
        Gemma-2 & 2B & 13.7 & 0.42 \\
        Gemma-2 & 9B & 14.4 & 1.10 \\
        \bottomrule
    \end{tabular}}
\end{table}

\section{Generalizability across LLM Architectures}
To verify that P2S is not dependent on specific models or parameter scales, we extended our evaluation to include various open-source LLMs: Llama-3.2-3B, Llama-3.1-8B, and Gemma-2 (2B/9B). As detailed in Table~\ref{tab:llm_generalization}, the performance variance across different models remains negligible ($<0.5\%$), despite significant differences in architecture and pre-training data. This confirms that our decomposition strategy relies on fundamental linguistic understanding rather than model-specific reasoning capabilities, ensuring the framework's robustness and flexibility.

\section{Additional Qualitative Analysis}
\label{sec:appendix_d}

\noindent \textbf{Qualitative Results on MomentSeeker.}
Figure~\ref{fig:cvpr_fig8} illustrates the qualitative result of P2S, demonstrating how our framework overcomes the dual challenges of Zero-shot Long Video Moment Retrieval (ZLVMR): semantic discrepancy and temporal search space saturation.

Figure~\ref{fig:cvpr_fig8} (a) highlights how P2S addresses semantic discrepancy during the \textit{refine} phase. The original query ($Q_o$: ``A man is holding a cup of coffee and walking towards a woman sitting on coach'') contains a compositional structure that single-vector embeddings often fail to capture accurately. To resolve this, our \textbf{Query Decomposition} module breaks the query into sub-queries: $Q_a$ (``holding a cup of coffee'') and $Q_b$ (``walking towards a woman''). These sub-queries serve as fine-grained evidence for Evidence-based Reranking, allowing the model to precisely align the visual segment [131s, 141s] with the semantic progression of the event, effectively eliminating the discrepancy between the complex query and the video representation.

Figure~\ref{fig:cvpr_fig8} (b) demonstrates the effectiveness of P2S in an hour-long video, addressing \textit{temporal search space saturation}. First, the \textbf{Adaptive Span Generator (ASG)} prevents candidate explosion by dynamically filtering out non-informative segments based on signal statistics. Subsequently, employing the \textbf{Evidence-based Reranking} and \textbf{Injection} mechanisms, the model utilizes the decomposed queries ($Q_a$: ``A soccer player in a game.'', $Q_b$: ``A soccer player falls to the ground during the game.'') as complementary evidence channels. By boosting candidates with strong peaks in both channels and injecting overlapping detections, the model precisely pinpoints the target event, ensuring robust localization even within a vast temporal search space.

\end{appendix}
